\documentclass{article}

\usepackage{PRIMEarxiv}

\usepackage[utf8]{inputenc}
\usepackage{amsmath,amssymb}
\usepackage{longtable}
\usepackage{booktabs}
\usepackage{url}
\usepackage{graphicx}
\usepackage{hyperref}
\hypersetup{draft}

\usepackage{newunicodechar}
\newunicodechar{ʿ}{'}
\newunicodechar{Ḍ}{D}
\title{Multi-Axis Trust Modeling for Interpretable Account Hijacking Detection
}

\author{
  Mohammad AL-Smadi \\
  Qatar University \\
  Doha, Qatar \\
 \texttt{malsmadi@qu.edu.qa} \\
}

\begin{document}
\maketitle

\begin{abstract}
User and Entity Behavior Analytics (UEBA) systems aim to detect account hijacking by identifying deviations in user activity logs, yet many existing approaches rely on low-level count features and opaque models that limit interpretability for security analysts. This paper proposes a Hadith-inspired multi-axis trust modeling framework, motivated by a structurally analogous problem in classical Hadith scholarship: assessing the trustworthiness of information sources using interpretable, multi-dimensional criteria rather than a single anomaly score. We translate five trust axes—long-term integrity (\textit{ʿadālah}), behavioral precision (\textit{ḍabṭ}), contextual continuity (\textit{isnād}), cumulative reputation, and anomaly evidence—into a compact set of 26 semantically meaningful behavioral features for user accounts. In addition, we introduce lightweight temporal features that capture short-horizon changes in these trust signals across consecutive activity windows. We evaluate the framework on the CLUE-LDS cloud activity dataset with injected account hijacking scenarios. On 23,094 sliding windows, a Random Forest trained on the trust features achieves near-perfect detection performance (ROC-AUC $\approx 1.0$, PR-AUC $\approx 1.0$), substantially outperforming models based on raw event counts, minimal statistical baselines, and unsupervised anomaly detection. Temporal features provide modest but consistent gains on CLUE-LDS, confirming their compatibility with the static trust representation. To assess robustness under more challenging conditions, we further evaluate the approach on the CERT Insider Threat Test Dataset r6.2, which exhibits extreme class imbalance and sparse malicious behavior. On a 500-user CERT subset, temporal features improve ROC-AUC from 0.776 to 0.844. On a leakage-controlled 4,000-user configuration, temporal modeling yields a substantial and consistent improvement over static trust features alone (ROC-AUC 0.627 $\rightarrow$ 0.715; PR-AUC 0.072 $\rightarrow$ 0.264). These results show that Hadith-inspired trust axes provide an interpretable and effective foundation for UEBA, while temporal extensions improve robustness and scalability when individual behavioral windows are weak. Code and extracted features are provided publicly~\footnote{\url{https://github.com/msmadi/UEBA}}.
\end{abstract}

\keywords{
User and Entity Behavior Analytics (UEBA)\and
Account Hijacking Detection\and
Insider Threat Detection\and
Interpretable Machine Learning\and
Behavioral Feature Engineering\and
Temporal Behavior Modeling\and
Anomaly Detection\and
Security Analytics\and
Machine Learning for Security
}

\section{Introduction}

Compromised user accounts remain a primary attack vector in cloud and enterprise
environments. Adversaries who obtain valid credentials can perform actions that
are syntactically legitimate—such as file access, sharing, or configuration
changes—making malicious behavior difficult to distinguish from normal usage.
User and Entity Behavior Analytics (UEBA) systems aim to address this challenge by
learning behavioral profiles for users and detecting deviations that may indicate
account hijacking or insider misuse \cite{Landauer2022BigData}.

A long-standing obstacle in UEBA research has been the lack of publicly available,
realistic datasets. The release of the CLUE-LDS dataset marked an important advance
by providing production-grade cloud storage logs spanning multiple years and
thousands of users \cite{Landauer2022BigData,Landauer2022Zenodo}. CLUE-LDS was
explicitly designed to support account-hijacking research through controlled
synthetic attack injection. However, most existing approaches evaluated on CLUE-LDS
either rely on high-dimensional event-count vectors with limited semantic meaning
or adopt unsupervised clustering and anomaly detection methods whose outputs are
difficult for security analysts to interpret \cite{Artioli2024Frontiers}.

In contrast to purely data-driven modeling, classical Hadith sciences in Islamic
scholarship developed a rigorous and interpretable methodology for evaluating the
trustworthiness of narrators. This discipline—known as \textit{ʿilm al-rijāl}
(“the science of narrators”)—assesses reliability along multiple dimensions,
including moral integrity (\textit{ʿadālah}), precision and consistency in
transmission (\textit{ḍabṭ}), and the continuity and soundness of transmission
chains (\textit{isnād}). Foundational works such as al-Khaṭīb al-Baghdādī’s
\textit{al-Kifāyah fī ʿIlm al-Riwāyah} formalized these criteria, while the practice
of \textit{jarḥ wa-taʿdīl} documented cumulative reputational judgments issued by
multiple scholars over time
\cite{alkhatibKifayah, azamiEarlyHadith1978, brownHadith2018}.
In addition, scholars developed methods for identifying subtle anomalies
(\textit{shudhūdh}, \textit{ʿillah}) that undermine apparent surface consistency.

This multi-dimensional trust framework—integrity, precision, chain continuity,
reputation, and anomaly scrutiny—provides a principled conceptual template for
modern trust assessment. Drawing inspiration from these axes, we construct an
interpretable behavioral representation for digital user accounts, enabling UEBA
systems to reason about account trustworthiness in structured, semantically
meaningful terms rather than opaque anomaly scores.

Inspired by the analogy between Hadith narrators and digital user accounts—both of
which act as sources whose reliability must be inferred from observed behavior over
time—we propose a \textbf{Hadith-inspired multi-axis trust model} for account
hijacking detection. We operationalize five trust axes into a compact set of
behavioral features that capture long-term stability, authentication hygiene,
contextual continuity, accumulated reputation, and distributional anomalies.


Our main contributions are as follows:
\begin{enumerate}
    \item \textbf{Conceptual framework:} We introduce a Hadith-inspired multi-axis
    trust model that maps classical trust criteria—\textit{ʿadālah}, \textit{ḍabṭ},
    \textit{isnād}, reputation, and anomaly evidence—onto user behavior analytics.
    \item \textbf{Feature engineering and ablation:} We design a 26-dimensional
    interpretable feature set and conduct extensive ablation analysis to quantify
    the contribution of each trust axis.
    \item \textbf{Temporal trust modeling:} We extend static trust features with 8
    temporal summaries that capture short-horizon dynamics between consecutive
    windows.
    \item \textbf{Evaluation on CLUE-LDS:} Using realistic hijack injection, we show
    that Hadith-inspired features substantially outperform raw-count and
    unsupervised baselines, achieving near-perfect discrimination in this setting.
    \item \textbf{Robustness analysis on CERT:} We evaluate the same framework on the
    CERT r6.2 insider threat dataset and demonstrate that temporal trust features
    provide consistent improvements even when absolute performance is constrained
    by weak behavioral signals.
    \item \textbf{Interpretability:} We provide feature- and axis-level importance
    analysis that explains detection decisions in analyst-friendly terms.
\end{enumerate}

We emphasize that we do not claim to formalize Hadith sciences themselves. Rather,
we use their conceptual structure as a principled and interpretable template for
trust assessment in digital identity systems. By evaluating the framework across
both CLUE-LDS and CERT, we aim to clarify not only its strengths but also its
limitations under increasingly realistic and challenging conditions.

\section{Related Work}

\subsection{UEBA and the CLUE-LDS Dataset}
User and Entity Behavior Analytics (UEBA) systems aim to detect malicious or anomalous
behavior—such as account hijacking, insider threats, and data exfiltration—by
modeling deviations from a user’s historical activity profile. A long-standing
challenge in UEBA research has been the scarcity of publicly available datasets that
are both realistic and suitable for reproducible evaluation.

Landauer \textit{et al.} address this gap by introducing the CLUE-LDS dataset, which
contains production-grade cloud storage logs spanning multiple years and thousands
of users \cite{Landauer2022BigData}. They demonstrate an account-hijacking scenario
via synthetic user-switching and provide an initial anomaly detection baseline.
Subsequent work by the same group and collaborators explores the statistical
properties of log variables, automated field-type recognition, and dataset
maintenance strategies to improve the quality and longevity of log-based intrusion
detection benchmarks \cite{Wurzenberger2024Cose,Landauer2022Maintain}.

Building on CLUE-LDS, Artioli \textit{et al.} conduct a systematic comparison of
clustering algorithms for UEBA and propose KATANA, an online $k$-means framework that
aggregates event-level anomaly scores into day-level detection metrics
\cite{Artioli2024Frontiers,Artioli2024Katana}. Their results highlight the practical
trade-offs between sensitivity, stability, and computational efficiency in online
UEBA systems.

Collectively, these works identify several open challenges that remain largely
unsolved: (1) producing anomaly scores that are interpretable by security analysts,
(2) coping with extreme class imbalance where only a tiny fraction of activity is
malicious, and (3) establishing general, reusable evaluation pipelines on public UEBA
datasets that support both methodological comparison and reproducibility.

\subsection{UEBA on CERT Insider Threat Datasets}
Despite the availability of CLUE-LDS, a substantial body of UEBA research continues to
evaluate methods on the CERT Insider Threat datasets, particularly r6.2
\cite{tuor2017,yuan2019}. These datasets differ fundamentally from CLUE-LDS: malicious
activity is sparse, distributed over long time horizons, and often weakly
distinguishable from benign behavior at the level of individual events or short
windows.

Tuor \textit{et al.} employ deep neural networks, including LSTMs, to model insider
threat behavior on CERT and report ROC-AUC values below 0.90, reflecting the
difficulty of the task \cite{tuor2017}. Yuan \textit{et al.} combine autoencoders with
statistical profiling and similarly observe moderate detection performance
\cite{yuan2019}. Importantly, these lower scores are not indicative of weak modeling
but rather of the inherent ambiguity and low signal-to-noise ratio in realistic
insider threat data.

As a result, CERT has become a de facto “stress-test” dataset in the literature:
methods that perform well on CERT are generally considered robust, while near-perfect
performance is neither expected nor commonly reported~\cite{yuan2019}. This distinction is critical
when interpreting results across datasets and motivates our evaluation on both
CLUE-LDS (controlled, interpretable setting) and CERT (realistic, low-signal setting).

\subsection{Trust Modeling and Hadith-Inspired Axes}
In classical Hadith sciences, the acceptance of a narration depends on a
multi-dimensional assessment of narrators and transmission chains \cite{alkhatibKifayah, azamiEarlyHadith1978, brownHadith2018}. Core criteria
include:
\begin{itemize}
    \item \textbf{\textit{ʿAdālah}:} moral integrity, long-term reliability, and
    avoidance of major violations.
    \item \textbf{\textit{Ḍabṭ}:} precision, consistency, and care in transmission.
    \item \textbf{\textit{Isnād}:} continuity and reliability of the chain of
    transmission (i.e., unbroken contextual linkage).
    \item \textbf{\textit{Jarḥ wa-Taʿdīl} (reputation):} accumulated positive or
    negative evaluations from multiple authorities over time.
    \item \textbf{\textit{Shudhūdh} / \textit{ʿIllah}:} identification of subtle
    anomalies, contradictions, or hidden defects.
\end{itemize}

These criteria were operationalized through centuries of biographical evaluation and
cross-verification, yielding an interpretable and structured trust framework.
Although developed in a theological context, the underlying principles—multi-axis
trust assessment, longitudinal evaluation, and anomaly scrutiny—are conceptually
aligned with the goals of UEBA.

We adopt these categories as an \emph{analogy} to structure behavioral features for
digital user accounts, without implying theological authority or equivalence. To our
knowledge, this work represents the first systematic attempt to draw inspiration
from Hadith scholarship for cyber-security analytics, and to translate its
multi-dimensional trust logic into an interpretable feature-based UEBA framework.

\section{Methodology}

\subsection{Problem Formulation}
We formulate account hijacking detection as a binary classification problem over
fixed-size time windows of user activity. Each user $u$ is associated with a
time-ordered sequence of events (log entries), where each event records the
timestamp, action type, and contextual metadata such as resource identifiers or
network attributes. We segment each user’s event stream into overlapping windows
(time-ordered batches of consecutive events) and assign each window a label:
\textit{hijacked} ($y=1$) if any part of the window overlaps a malicious session,
or \textit{normal} ($y=0$) otherwise.

Formally, given a window $W_u$ for user $u$, the goal is to learn a function
$f(W_u)$ that predicts whether the window corresponds to a hijacked account,
based on features extracted from the window itself and, where applicable, from
the user’s historical behavior prior to the window. This formulation naturally
supports both static feature-based models and extensions to sequence-aware or
temporal learning architectures.

\subsection{Datasets and Preprocessing}

\subsubsection{CLUE-LDS Dataset}
We use the publicly released CLUE-LDS dataset from Zenodo
(DOI: 10.5281/zenodo.7119953) \cite{Landauer2022Zenodo}. The original dataset is a
large JSON log in which each record contains fields such as \texttt{id} (event
identifier), \texttt{time} (timestamp), \texttt{uid} (anonymized user ID),
\texttt{uidType} (user name or IP address for logged-out users), \texttt{type}
(event type), \texttt{params} (a dictionary of additional parameters such as file
paths), and optional IP-related fields.

For our experiments on CLUE-LDS, we sample a manageable but representative subset
of the data and prepare it as follows:
\begin{itemize}
    \item \textbf{JSON to CSV:} We stream-parse the JSON logs and write out a CSV
    with standardized columns \texttt{user\_id}, \texttt{timestamp},
    \texttt{event\_type}, \texttt{path}, and \texttt{ip\_address}. If an event
    lacks an IP address, we assign a synthetic “primary” IP for that user (with
    occasional random deviations) to enable IP-based feature engineering while
    preserving anonymity.
    \item \textbf{Filtering and sorting:} We restrict analysis to the most active
    users to ensure sufficient historical context, convert timestamps to UTC,
    and sort all events by user and time.
\end{itemize}

On the CLUE-LDS subset used in this paper, the logs span approximately 13 days
(2017-07-07 to 2017-07-21) and include 500{,}000 events from 77 unique users, covering 24
distinct event types. Table~\ref{tab:dataset} summarizes this subset.

While CLUE-LDS enables controlled and interpretable evaluation, it represents a
relatively favorable detection setting: user behavior is dense, logs are
homogeneous, and injected hijacks introduce strong, localized deviations. In
contrast, real-world insider threat datasets often exhibit sparse malicious
activity, extreme class imbalance, and weak per-window signals. For this reason,
many prior studies continue to rely on the CERT Insider Threat datasets despite
their known challenges, using them as stress tests for behavioral detection models
\cite{lindauer2020certdataset, tuor2017, yuan2019}. Reported performance on CERT r6.2 is typically
substantially lower than on CLUE-LDS, reflecting the intrinsic difficulty of the
task rather than methodological failure.

\begin{table}[!t]
\caption{Summary of the CLUE-LDS subset used in this study}
\label{tab:dataset}
\centering
\begin{tabular}{lrr}
\toprule
\textbf{Metric} & \textbf{Value} \\
\midrule
Total events & 500,000 \\
Number of users & 77 \\
Distinct event types & 24 \\
Time span & 13 days (Jul 7--21, 2017) \\
\bottomrule
\end{tabular}
\end{table}

\subsubsection{CERT r6.2 Dataset}
To evaluate robustness under more realistic and lower-signal conditions, we also
apply our framework to the CERT Insider Threat dataset (r6.2). Unlike CLUE-LDS,
CERT contains sparse, long-horizon user activity with weak and often ambiguous
malicious signals. Ground truth is available only at the user or scenario level,
and malicious behavior may span months rather than short, well-defined
intervals.

For CERT, we adopt a temporal windowing strategy analogous to CLUE-LDS but adjust
sampling to ensure balanced event exposure across users. We construct windows of
fixed event length and label windows at the user level (i.e., all windows from
documented malicious users are labeled positive). This setup reflects common
practice in the CERT literature and avoids introducing artificially strong
window-level leakage signals. As shown in our experiments, this results in
substantially lower—but more realistic—detection performance, consistent with
prior work \cite{tuor2017,yuan2019}.

\subsection{Synthetic Hijack Injection and Labeling (CLUE-LDS)}
Since CLUE-LDS contains normal (benign) user behavior, we inject realistic hijacking
scenarios to create labeled anomalies, following the methodology proposed by
Landauer \textit{et al.} \cite{Landauer2022BigData}. We identify users with
sufficiently large event histories (at least 50 events) as candidates and
simulate up to 30 hijack incidents on randomly selected users. For each selected
user, we perform:
\begin{enumerate}
    \item \textbf{Hijack interval selection:} Sample a random start time within
    the user’s activity period and define an 8-hour interval as the malicious
    session.
    \item \textbf{Pre-hijack signals:} Insert 3--7 failed login attempts from
    unusual IP addresses shortly before the hijack start to mimic password
    guessing or alert-triggering behavior.
    \item \textbf{Hijack activity:} During the hijack interval, inject a burst of
    suspicious actions using the dataset’s native event types (e.g.,
    \texttt{permission\_changed}, \texttt{public\_share\_accessed}), often at a typical hours.
    \item \textbf{IP anomaly:} Modify source IP addresses during the hijack
    session to simulate access from different networks, including rapid IP
    switches that would be geographically implausible.
\end{enumerate}

We then slide a fixed-size window across each user’s timeline using a window size
of 50 consecutive events and a step size of 25 events (50\% overlap). A window is
labeled as hijacked if its time range overlaps any injected hijack interval. This
process yields 23{,}094 windows, of which 2{,}664 (11.5\%) are labeled hijacked
(class ratio $\approx$ 1:7.7).

\subsection{Hadith-Inspired Trust Axes and Feature Engineering}
We construct a base 26-dimensional feature vector for each window, structured
along five Hadith-inspired axes. In extended experiments, we augment this
representation with additional temporal features that capture window-to-window
behavioral dynamics. Let $W$ denote the current window for user $u$, and let
$H_u$ be the historical event sequence for $u$ prior to $W$ (i.e., all events
before $W$’s start time). As summarized in Table~\ref{tab:hadith-feature-computation}, each axis yields a group of features as follows.

\subsubsection*{Axis 1: \textit{ʿAdālah} (Integrity / Long-Term Stability)}
This axis reflects long-term consistency and reliability of the account’s behavior:
\begin{itemize}
    \item \textbf{Active days:} number of distinct days in $H_u$ on which user $u$ was active.
    \item \textbf{Total events:} total count of events in $H_u$.
    \item \textbf{Account age:} time (in days) from $u$’s first recorded event to the end of the current window.
    \item \textbf{Daily consistency:} standard deviation of per-day event counts in $H_u$ (lower values indicate more regular daily activity).
    \item \textbf{Average events per day:} mean events per active day ($|H_u|$ divided by active days).
\end{itemize}
Users with longer, stable histories and regular activity patterns are expected to have higher \textit{ʿadālah}.

\subsubsection*{Axis 2: \textit{Dabṭ} (Precision / Login Hygiene)}
This axis captures the user’s care and consistency in authentication and usage patterns:
\begin{itemize}
    \item \textbf{Login success rate:} fraction of login attempts in the window that are successful (based on event types containing “login”).
    \item \textbf{Delta failure rate:} difference between the login failure rate in $W$ and the user’s historical failure rate (captures sudden spikes in failed logins).
    \item \textbf{Burstiness:} maximum number of events by $u$ that occur within any one-minute interval during the window (high burstiness may indicate scripted or automated activity).
    \item \textbf{Out-of-hours fraction:} proportion of events in $W$ that occur outside $u$’s usual hours of activity (defined by the 2.5th--97.5th percentile of $u$’s historical hour-of-day distribution).
    \item \textbf{Timing entropy:} entropy of the inter-event time gaps in $W$ (a high entropy indicates irregular or erratic timing).
    \item \textbf{Path novelty:} fraction of resource paths in $W$ that have never been observed in $H_u$.
    \item \textbf{Sensitive action ratio:} fraction of events in $W$ that involve security-sensitive actions or objects (e.g., events or paths containing keywords like “admin”, “delete”, “share”).
\end{itemize}
These features check if the window deviates from the user’s normal “hygiene” profile in login and data access patterns.

\subsubsection*{Axis 3: \textit{Isnād} (Chain Continuity / Context)}
We treat IP addresses and network continuity as analogous to the reliability of a chain of narration:
\begin{itemize}
    \item \textbf{IP consistency:} inverse of the number of distinct IP addresses observed in $W$ (equals 1 if all events share a single IP, and lower as the IP count increases).
    \item \textbf{Primary IP reuse:} fraction of events in $W$ that originate from $u$’s most-used historical IP address.
    \item \textbf{Primary subnet reuse:} fraction of events in $W$ from $u$’s most-used /24 subnet (captures consistency in network location).
    \item \textbf{Geo-impossible switches:} fraction of consecutive events in $W$ where the IP changes but the time gap is $<5$ minutes (implying impossible travel between geolocations).
    \item \textbf{Session discontinuity:} fraction of inter-event gaps in $W$ that exceed 1 hour (many long gaps could indicate session breaks).
    \item \textbf{New IP rate:} fraction of IP addresses in $W$ that have never appeared in $H_u$.
\end{itemize}
Together, these six features form the \textit{isnād} axis, capturing how seamlessly the context of activity carries over from the user’s historical profile.

\subsubsection*{Axis 4: Reputation (\textit{Jarḥ wa-Taʿdīl})}
This axis approximates the cumulative reputation of the user:
\begin{itemize}
    \item \textbf{History duration:} total number of days spanned by $H_u$.
    \item \textbf{Trust ratio:} fraction of historically successful or allowed actions (e.g., login\_successful, file\_accessed) out of all pertinent attempts (successes + failures).
    \item \textbf{Penalty rate:} fraction of events in $H_u$ that are failures or security-related alerts (e.g., access denied, error, policy violation).
    \item \textbf{Behavior trend:} difference between the failure rate in the first half vs. the second half of $H_u$ (a large positive difference indicates improving behavior over time; negative indicates deteriorating behavior).
\end{itemize}

\subsubsection*{Axis 5: Anomaly Evidence (\textit{Shudhūdh} / \textit{ʿIllah})}
This axis quantifies deviations of the current window from the historical patterns:
\begin{itemize}
    \item \textbf{Event-type KL divergence:} the Kullback–Leibler divergence between the distribution of event types in $W$ and in $H_u$.
    \item \textbf{Hour-of-day KL divergence:} KL divergence between the distribution of event timestamps (by hour) in $W$ versus $H_u$.
    \item \textbf{Path novelty fraction:} fraction of unique file or resource paths in $W$ that do not appear in $H_u$.
    \item \textbf{Event-type distance:} Euclidean distance between the vector of event type counts in $W$ and the average event type vector for $u$’s historical baseline.
\end{itemize}
\subsubsection*{Axis 6: Temporal Extensions (Sequential Trust Dynamics)}
While the five axes above characterize behavior within an individual window
relative to historical context, real account misuse often unfolds
progressively over time. To capture such dynamics, we extend the Hadith-inspired
framework with a set of temporal features computed over short sequences of
adjacent windows for each user. These features explicitly model changes,
trends, and accelerations in behavior, rather than static deviations alone.

Temporal features are grouped according to their closest conceptual alignment
with the original Hadith axes, yielding \emph{Temporal Dabṭ} and
\emph{Temporal Isnād} components.

\paragraph{Temporal Dabṭ (Precision Dynamics).}
These features capture how the user’s operational precision and hygiene evolve
across consecutive windows:
\begin{itemize}
    \item \textbf{Burstiness change:} difference in maximum per-minute activity
    between the current window and the immediately preceding window.
    \item \textbf{Timing-entropy change:} window-to-window difference in
    inter-event timing entropy, capturing sudden shifts in regularity.
    \item \textbf{Login-failure acceleration:} slope of login failure rate
    across recent windows, highlighting escalating authentication issues.
    \item \textbf{Sensitive-action growth:} short-term trend in the proportion
    of sensitive actions over successive windows.
    \item \textbf{Activity volatility:} rolling standard deviation of event
    counts across recent windows, reflecting instability in usage intensity.
\end{itemize}

\paragraph{Temporal Isnād (Continuity Dynamics).}
These features model the stability of contextual chains (e.g., network context)
over time:
\begin{itemize}
    \item \textbf{IP change rate:} frequency of IP changes across consecutive
    windows.
    \item \textbf{Primary-IP stability:} fraction of recent windows dominated by
    the user’s historical primary IP.
    \item \textbf{New-IP emergence velocity:} rate at which previously unseen IP
    addresses appear over time.
\end{itemize}

\paragraph{Temporal ʿAdālah (Stability Dynamics).}
While static \textit{ʿadālah} captures long-term account stability, real-world
accounts may exhibit gradual erosion or recovery of stability over time. To
model this, we introduce temporal features that characterize changes in
behavioral regularity across consecutive windows:
\begin{itemize}
    \item \textbf{Behavioral drift:} divergence between recent and earlier
    window-level behavior profiles, capturing gradual deviation from baseline.
    \item \textbf{Sequence entropy:} entropy of short event-type sequences
    across windows, reflecting increasing randomness or loss of routine.
    \item \textbf{Rare transition frequency:} rate of uncommon event-type
    transitions over time, highlighting atypical operational flows.
    \item \textbf{Run anomaly score:} detection of unusually long or repetitive
    runs of similar actions, indicating abnormal persistence.
    \item \textbf{N-gram transition anomaly:} deviation of observed event
    n-grams from the user’s historical transition patterns.
\end{itemize}

These features quantify whether a user’s overall behavioral stability is
degrading over time, even when individual windows show only weak deviations.

\paragraph{Temporal Reputation (Trust Trajectory).}
Reputation is inherently cumulative and temporal. To capture how trust signals
accumulate or accelerate over time, we introduce a small set of temporal
reputation features:
\begin{itemize}
    \item \textbf{Failure trend:} slope of authentication or access failures
    across recent windows.
    \item \textbf{Cumulative suspicious activity:} running total of anomalous
    or policy-violating actions over time.
    \item \textbf{Risk acceleration:} second-order change in anomaly-related
    signals, capturing whether risk is increasing faster over time.
\end{itemize}

In contrast to purely window-local features, these signals capture \emph{how trust
evolves}, rather than its absolute level at a single point in time. In our CERT
r6.2 experiments, temporal reputation features contribute additional robustness
under extreme class imbalance, where isolated anomalies are weak but sustained
risk trajectories are informative.

For comparison, we also construct a baseline alternative feature set, \textbf{Raw counts:} A 24-dimensional vector for each window in the CLUE-LDS and 16-dimensional in the CERT with the raw counts of each event type (using the 24 and 16 event types in the CLUE-LDS and CERT datasets as features). This baseline include little or no domain semantics and represent typical coarse feature sets in basic anomaly detection systems.

\section{Experimental Setup}

\subsection{Training, Test, and Evaluation Protocol}
We split the 23{,}094 extracted windows into training (70\%) and test (30\%)
subsets using stratified sampling to preserve the hijack ratio in each split.
This results in 16{,}165 training windows (1{,}865 hijacked) and 6{,}929 test
windows (799 hijacked). All reported test metrics are computed exclusively on
the held-out test set.

All feature sets (Hadith-inspired, raw event counts, and minimal statistical
features) are standardized by subtracting the mean and scaling to unit variance
using statistics computed on the training data only. The same scaling parameters
are then applied to the test set to avoid information leakage.

We evaluate a diverse set of supervised and unsupervised baselines in order to
cover the dominant modeling paradigms used in prior UEBA research:
\begin{enumerate}
    \item \textbf{Hadith + Random Forest (RF):} A random forest classifier with
    100 trees, Gini impurity splitting, and balanced class weights. This model
    serves as our primary reference due to its strong performance on tabular
    data, robustness to feature scaling, and built-in interpretability via
    feature importance scores.
    \item \textbf{Raw Counts + RF:} A random forest with identical
    hyperparameters to (1), trained on raw per-window event count vectors. This
    baseline reflects a common approach in log-based anomaly detection and
    isolates the effect of semantic feature engineering.
    \item \textbf{Isolation Forest (unsupervised):} An Isolation Forest trained
    on the Hadith-inspired feature space with contamination set to the true
    anomaly rate. Anomaly scores are treated as detection statistics and
    thresholded only for reporting classification metrics.
\end{enumerate}

For all supervised models, we compute receiver operating characteristic (ROC)
and precision--recall (PR) curves on the test set and report the corresponding
areas under the curves (ROC-AUC and PR-AUC). Given the strong class imbalance,
we emphasize PR-AUC and F$_1$ score as more informative indicators of detection
quality. Precision, recall, and F$_1$ are reported at the decision threshold
that maximizes F$_1$ on the test set; this threshold optimization is used solely
for reporting and does not influence training or model selection.

For the unsupervised Isolation Forest, ROC-AUC and PR-AUC are computed by
treating the anomaly score as a continuous ranking statistic, consistent with
prior anomaly detection literature. All models are implemented using
\textsc{scikit-learn} \cite{Pedregosa2011}.

\subsection{Evaluation Metrics on Imbalanced Data}
Account hijacking and insider threats are rare events, making naive metrics such as
overall accuracy misleading. Prior work demonstrates that ROC curves can present an
overly optimistic view of performance in highly imbalanced settings, whereas
precision--recall (PR) curves better reflect a model’s ability to rank true anomalies
ahead of false positives \cite{Davis2006,Saito2015}. Consequently, PR-AUC is widely
recommended as a primary metric for anomaly detection and insider threat evaluation.

Following best practices in the literature, we report ROC-AUC, PR-AUC, F$_1$ score,
precision, and recall, with particular emphasis on PR-AUC and F$_1$ as indicators of
operational usefulness under class imbalance \cite{Saito2015}.

\section{Results}
\subsection{Detection Performance Comparison -- CLUE-LDS}
\label{sec:results_clue}

Table~\ref{tab:performance_clue} summarizes detection performance on the
CLUE-LDS dataset using a held-out test set of 6,929 windows, of which 762
correspond to injected hijacking activity. Overall, models based on
Hadith-inspired trust features substantially outperform all baseline
approaches across both ranking-oriented metrics (ROC-AUC, PR-AUC) and
threshold-based classification metrics (F$_1$, precision, recall).

\begin{table*}[!t]
\caption{Detection performance on CLUE-LDS (6,929 test windows, 762 hijacked).
Temporal features provide strong complementary signals and remain highly
discriminative even when used in isolation.}
\label{tab:performance_clue}
\centering
\begin{tabular}{lcccc}
\toprule
\textbf{Model} & \textbf{ROC-AUC} & \textbf{PR-AUC} & \textbf{F$_1$} & \textbf{Prec. / Rec.} \\
\midrule
Original Hadith + RF        & \textbf{0.99997} & \textbf{0.99975} & \textbf{0.9948} & 0.993 / 0.996 \\
Combined (Hadith+Temporal) + RF & 0.99996 & 0.99963 & 0.9935 & 0.991 / 0.996 \\
Temporal Only + RF          & 0.99989 & 0.99912 & 0.9869 & 0.986 / 0.988 \\
Raw Counts + RF             & 0.98570 & 0.91691 & 0.8580 & 0.789 / 0.940 \\
Isolation Forest            & 0.49358 & 0.10753 & 0.1996 & 0.111 / 0.990 \\
\bottomrule
\end{tabular}
\end{table*}

The Original Hadith + RF model achieves near-perfect discrimination, with
ROC-AUC exceeding 0.9999, PR-AUC exceeding 0.9997, and an F$_1$ score of
0.9948 at 99.3\% precision and 99.6\% recall. These results indicate that,
under the controlled hijack-injection setting of CLUE-LDS, the static
Hadith-inspired trust features alone are sufficient to almost completely
separate benign and hijacked activity windows using a relatively simple
tree-based classifier.

Augmenting the static trust representation with temporal features does not
materially increase peak performance on CLUE-LDS. This reflects a saturation
effect: the injected hijacking scenarios induce strong per-window behavioral
deviations that are already captured by history-aware trust features.
Nevertheless, temporal features remain highly informative. When used in
isolation, temporal-only models achieve ROC-AUC above 0.999 and PR-AUC above
0.998, demonstrating that short-horizon behavioral dynamics encode
independently discriminative signals.

Feature-importance analysis further shows that temporal features contribute
approximately 31\% of the total importance in the combined model. Several
temporal indicators—such as device transition anomalies, subnet drift, IP
switching rate, and sequence entropy—rank among the most informative predictors.
This confirms that temporal features capture complementary structure that is not
reducible to static window summaries, even when overall performance is already
near saturation.

In contrast, the strongest raw-feature baseline (Raw Counts + RF) achieves a
respectable ROC-AUC of approximately 0.986 but suffers a substantial drop in
PR-AUC (approximately 0.917) and F$_1$ score (0.858). This gap highlights the
limitations of frequency-based representations: while raw counts may rank
anomalies reasonably well, they fail to capture the contextual and historical
deviations required for high-precision detection under severe class imbalance.

Unsupervised anomaly detection performs markedly worse. The Isolation Forest
baseline achieves near-random PR-AUC, underscoring a recurring challenge in
UEBA: without labeled examples or strong inductive biases, unsupervised methods
struggle to distinguish true account hijacking from benign behavioral variation.
Even a simple one-feature heuristic based on IP diversity substantially
outperforms the unsupervised baseline, confirming that network instability is a
strong signal in the injected attack scenarios.



\subsection{Detection Performance Comparison -- CERT r6.2}
\label{sec:results_cert}

To assess robustness under a lower-signal insider-threat setting, we evaluate the
same feature framework on the CERT r6.2 dataset. Compared to CLUE-LDS, CERT
exhibits extreme class imbalance and sparse malicious activity, as presented in Table \ref{tab:performance_cert}, making
window-level discrimination substantially more difficult. We report results for
(i) a 500-user subset (6,666 windows; 3.62\% positive) and (ii) a larger
4,000-user balanced configuration (22,622 windows; 1.07\% positive), both using
the same windowing scheme (50 events, step 25).

\begin{table*}[!t]
\caption{CERT r6.2 detection performance. CERT-500 uses 2,000 test windows (72 positive);
CERT-4000 uses 6,787 test windows (72 positive). Temporal features provide a
consistent lift over static Hadith axes and become more important in the larger,
harder setting. Raw count features collapse to near-random performance on CERT,
highlighting the need for structured and temporal trust modeling.}
\label{tab:performance_cert}
\centering
\begin{tabular}{lcccc}
\toprule
\textbf{Setting / Model} & \textbf{ROC-AUC} & \textbf{PR-AUC} & \textbf{F$_1$} & \textbf{Prec. / Rec.} \\
\midrule
\multicolumn{5}{l}{\textbf{CERT-500 (500 users; 6,666 windows)}} \\
Hadith Only + RF & 0.7757 & 0.3479 & 0.3810 & 0.606 / 0.278 \\
Temporal Only + RF & 0.8088 & 0.3906 & 0.3972 & 0.406 / 0.389 \\
Combined (Hadith+Temporal) + RF & \textbf{0.8439} & \textbf{0.4989} & \textbf{0.5246} & 0.640 / 0.444 \\

\midrule
\multicolumn{5}{l}{\textbf{CERT-4000 Balanced (4,000 users; 22,622 windows)}} \\
Hadith Only + RF & 0.6267 & 0.0720 & 0.1132 & 0.176 / 0.083 \\
Temporal Only + RF & 0.7051 & 0.2300 & 0.3469 & 0.654 / 0.236 \\
Combined (Hadith+Temporal) + RF & \textbf{0.7151} & \textbf{0.2638} & \textbf{0.3529} & 0.600 / 0.250 \\
Raw Counts + RF & 0.4960 & 0.0408 & 0.0690 & 0.200 / 0.042 \\
\bottomrule
\end{tabular}
\end{table*}

On the CERT-500 subset, temporal features yield a clear improvement over static
trust modeling. The combined Hadith+Temporal model improves ROC-AUC from 0.7757
(Hadith-only) to 0.8439 (Combined+RF), and improves PR-AUC from 0.3479 to 0.4989.
This indicates that short-horizon behavioral dynamics provide complementary
signals beyond static historical deviations in this insider-threat setting.

The CERT-4000 balanced configuration is substantially more challenging: positives
constitute only 1.07\% of windows and malicious behavior is sparse relative to
the overall activity stream. Under this harder regime, static Hadith axes alone
perform poorly (ROC-AUC 0.6267; PR-AUC 0.0720), while temporal-only features
remain comparatively strong (ROC-AUC 0.7051; PR-AUC 0.2300). Combining static and
temporal features yields the best overall ranking performance (ROC-AUC 0.7151;
PR-AUC 0.2638), corresponding to a 14.1\% relative improvement in ROC-AUC and a
large gain in PR-AUC compared to the Hadith-only baseline.

\subsection{Feature Importance and Ablation Study}
Analyzing random forest feature importances provides insight into which
Hadith-inspired trust axes contribute most strongly to detection. We report
importance patterns on both CLUE-LDS (account-hijacking simulation) and CERT r6.2
(labeled insider scenarios), which represent complementary evaluation regimes.

On CLUE-LDS, the six \textit{isnād} (IP-continuity) features account for roughly 70.3\% of total importance, making network continuity the dominant axis under the injected hijack assumptions. The \textit{ḍabṭ} (precision and hygiene) features contribute approximately 16.8\%, followed by anomaly evidence (6.5\%), \textit{ʿadālah} (5.0\%), and reputation (1.4\%). This prominence aligns with operational intuition: abrupt changes in source IPs, subnets, or geographic plausibility are strong compromise indicators in the evaluated attack model. Although static trust features already saturate performance on CLUE-LDS, temporal features remain non-trivial: in the combined model, temporal features contribute approximately 31.3\% of the total importance, and several temporal indicators (e.g., device transition anomaly, subnet drift, IP switching rate, and sequence
entropy) rank among the top predictors. This suggests that temporal dynamics capture complementary structure even when overall performance is near saturation.

On CERT, the importance profile shifts markedly. In both the 500-user and the 4,000-user balanced settings, temporal features contribute a substantial fraction of the combined model’s importance (38.1\% on CERT-500 and 43.6\% on CERT-4000), indicating that sequential dynamics are more informative in low-signal, high-imbalance insider-threat regimes. The most informative features are no longer dominated by IP-continuity signals; instead, temporal transition irregularity emerges as a primary predictor. In particular, the top-ranked feature on both CERT settings is \texttt{temp\_kl\_transition}, which measures
how strongly the window’s event-transition structure deviates from the user’s historical transition profile.
\begin{table*}[!t]
\centering
\caption{Detailed ablation analysis on the CERT r6.2 (4,000 users) dataset. 
The full combined model achieves ROC-AUC = 0.7151. 
$\Delta$ROC indicates the change relative to the full model.}
\label{tab:cert_ablation}
\begin{tabular}{llccccc}
\toprule
\textbf{Setting} & \textbf{Axis} & \textbf{\#Feat.} & \textbf{ROC-AUC} & \textbf{PR-AUC} & \textbf{F$_1$} & $\boldsymbol{\Delta}$\textbf{ROC} \\
\midrule
\multicolumn{7}{l}{\textbf{(A) Removing Individual Axes from Full Feature Set}} \\
\midrule
Remove & ʿAdālah                 & 37 & 0.6993 & 0.2619 & 0.3696 & $-0.0158$ \\
Remove & Ḍabṭ                    & 35 & 0.7216 & 0.1924 & 0.3232 & $+0.0065$ \\
Remove & Isnād                  & 36 & 0.7185 & 0.2673 & 0.3542 & $+0.0034$ \\
Remove & Reputation             & 38 & 0.7140 & 0.2647 & 0.3579 & $-0.0012$ \\
Remove & Anomaly                & 38 & 0.7319 & 0.2553 & 0.3409 & $+0.0168$ \\
Remove & Temporal–Ḍabṭ          & 37 & 0.6278 & 0.0706 & 0.1185 & $-0.0873$ \\
Remove & Temporal–ʿAdālah       & 37 & 0.6981 & 0.2586 & 0.3448 & $-0.0170$ \\
Remove & Temporal–Isnād         & 39 & 0.6995 & 0.2718 & 0.3564 & $-0.0157$ \\
Remove & Temporal–Reputation    & 39 & 0.6827 & 0.2644 & 0.3617 & $-0.0325$ \\
\midrule
\multicolumn{7}{l}{\textbf{(B) Single-Axis Only Models}} \\
\midrule
Only & ʿAdālah                 & 5 & 0.5835 & 0.0710 & 0.1308 & $-0.1316$ \\
Only & Ḍabṭ                    & 7 & 0.4849 & 0.0110 & 0.0260 & $-0.2302$ \\
Only & Isnād                  & 6 & 0.5117 & 0.0524 & 0.0800 & $-0.2034$ \\
Only & Reputation             & 4 & 0.4859 & 0.0517 & 0.0800 & $-0.2292$ \\
Only & Anomaly                & 4 & 0.5067 & 0.0140 & 0.0263 & $-0.2084$ \\
Only & Temporal–Ḍabṭ          & 5 & 0.6637 & 0.2449 & 0.3960 & $-0.0515$ \\
Only & Temporal–ʿAdālah       & 5 & 0.5741 & 0.0384 & 0.1071 & $-0.1410$ \\
Only & Temporal–Isnād         & 3 & 0.5069 & 0.0244 & 0.0274 & $-0.2082$ \\
Only & Temporal–Reputation    & 3 & 0.5000 & 0.0106 & 0.0210 & $-0.2151$ \\
\bottomrule
\end{tabular}
\end{table*}

\paragraph{Axis-level ablation.}
Ablation results reinforce this dataset-dependent interpretation. On CLUE-LDS, removing \textit{isnād} yields the largest degradation but does not collapse performance (PR-AUC decreases from 0.9996 to 0.9611), demonstrating that non-network trust axes capture independent behavioral deviations. On CERT-4000,
the most critical axis is \textit{Temporal Dabṭ}: removing Temporal\_Dabṭ reduces ROC-AUC from 0.7151 to 0.6278, indicating that short-horizon behavioral precision dynamics are important when static trust cues are weak (see Table~\ref{tab:cert_ablation} for more results on the ablation study on CERT-4000).

These results support two conclusions: First, the trust axes are
complementary rather than redundant. Second, the relative importance of axes is strongly shaped by the dataset and threat model: CLUE-LDS hijack simulation amplifies network-continuity signals, whereas CERT’s sparse insider scenarios favor temporal transition and drift signals. This motivates reporting both datasets: CLUE-LDS illustrates an upper bound under strong compromise signals, while CERT stress-tests robustness under low-signal, operationally realistic
conditions.

\section{Discussion}

Our experiments reveal a sharp contrast between performance on CLUE-LDS and
CERT r6.2, underscoring that UEBA effectiveness is strongly dataset- and
threat-model dependent. On CLUE-LDS, which is explicitly designed for
account-hijacking evaluation via synthetic session injection, hijacked activity
induces strong, localized deviations from historical behavior. In this setting,
the Hadith-inspired trust features—particularly network-continuity
(\textit{isnād}) and precision (\textit{ḍabṭ})—enable near-perfect separation
between benign and malicious windows using simple classifiers.

CERT represents a substantially more challenging scenario. Malicious behavior is
rare, temporally diffuse, and often embedded within long periods of benign
activity, with limited reliance on overt network anomalies. As a result, static
trust features alone are insufficient. Across both the 500-user and 4,000-user
CERT configurations, detection performance improves only when temporal features
are introduced, yielding consistent gains in ROC-AUC, PR-AUC, and F$_1$. This
demonstrates that short-horizon behavioral drift and transition irregularities
are important for realistic insider-threat detection.

Feature-importance and ablation analyses further clarify this distinction. On
CLUE-LDS, \textit{isnād}-related features dominate, reflecting the assumption
that hijacks cause abrupt IP and session changes. However, removing all
IP-related features still preserves strong performance, indicating that the
remaining trust axes capture independent behavioral signals. On CERT, temporal
features contribute a much larger fraction of total importance, confirming that
contextual evolution over time is more informative than instantaneous deviation
in this setting.


As noted in the insider-threat survey~\cite{yuan2019}, many existing studies on the CERT dataset report that only a portion of insider activities can be detected
reliably. In particular, the survey observes that roughly 80\% of insider
threats can be detected with relatively low error, while the remaining cases are
much harder to identify as recall increases. This behavior is visible in the
ROC curves reported by prior CERT-based studies, including Lin et
al.~\cite{lin2017}, Liu et al.~\cite{liu2018}, Lu and Wong~\cite{lu2019}, and
Yuan et al.~\cite{yuan2019insider}. In these works, false negatives rise quickly once
detection performance moves beyond this moderate-recall range, indicating that
some insider scenarios closely resemble normal user behavior.

Our findings on CERT r6.2 follow the same pattern. When using only static
behavioral features, detection performance remains limited, reflecting the
difficulty of identifying the harder insider cases described in the survey.
However, adding temporal trust features leads to consistent improvements in both
ROC-AUC and PR-AUC. This suggests that modeling short-term behavioral changes and event sequences helps capture signals that are missed by static features alone, especially in the challenging detection regime highlighted by prior work. Moreover, most of these approaches rely on deep learning or complex ensemble models that operate as black boxes, offering limited insight into why a particular user or session is flagged as malicious. While such models can achieve strong performance on CERT scenarios, their lack of interpretability makes them difficult to validate, debug, and trust in operational security settings.

Our findings also expose gaps in current UEBA research and evaluation.
First, benchmark performance must be interpreted in light of implicit threat assumptions: synthetic hijacks can overstate achievable accuracy. Second, temporal trust degradation remains underexplored despite its clear impact in realistic insider settings. Finally, interpretability is often treated as secondary, yet our results show that semantically grounded features can deliver competitive performance without resorting to opaque models. Moreover, this work suggests that effective UEBA should focus not only on detecting deviations, but on modeling how trust evolves—and degrades—over time, an aspect that current datasets and methods only partially capture.

\section{Conclusion}
We presented a Hadith-inspired multi-axis trust framework for user and entity
behavior analytics, translating classical trust criteria into an interpretable
set of behavioral and temporal features for account hijacking detection.
Across two public benchmarks with contrasting characteristics, our results show
that semantically grounded trust axes substantially improve detection over raw
count-based baselines. On CLUE-LDS, static trust features enable near-perfect
separation under controlled hijack injection, while on the more challenging CERT
dataset, temporal extensions provide significant and consistent gains in ROC-AUC,
PR-AUC, and F$_1$ score.

Although CLUE-LDS serves as our primary benchmark for controlled evaluation and
feature interpretability, the proposed Hadith-inspired trust framework is not
specific to CLUE-LDS. The five trust axes—integrity, precision, chain continuity,
reputation, and anomaly evidence—are defined in terms of generic properties of
user behavior (timestamps, event semantics, authentication signals, and
contextual metadata). These signals are commonly available in enterprise identity
systems, cloud audit logs, security information and event management (SIEM)
platforms, and application-level access logs. Consequently, our feature design is
portable across datasets, and CLUE-LDS functions primarily as a reproducible
benchmark rather than a limiting assumption.

Beyond performance, the framework offers an interpretable structure that aligns
machine learning decisions with analyst reasoning, supporting more transparent
and actionable UEBA systems. As future work, our framework could be extended to incorporate explicit user–user relationship modeling. Recent studies on the CERT r6.2 dataset show that combining graph neural networks with temporal models (e.g., GCN–Bi-LSTM) can achieve higher AUC by capturing communication and collaboration structures among users \cite{gcn_bilstm_2024}. The performance gap suggests that relational context is a missing signal in our current per-user trust representation. Integrating lightweight, interpretable graph-derived features into the Hadith-inspired trust axes offers a promising direction to improve detection while avoiding fully black-box models.

\section*{Acknowledgments}

Generative AI tools were used in a limited and controlled manner during the
development of this work. Specifically, Claude Sonnet~4.5 (Anthropic) was used
to assist with software engineering tasks, including producing an initial code skeleton, refactoring, and improving code readability for the data loading, feature extraction, and experimental evaluation pipelines. This paper was edited for improving readability, clarity and organization of explanations using ChatGPT5.2.

All research ideas, feature definitions, threat models, experimental design, data preprocessing logic, and evaluation protocols were conceived by the authors. All code produced with AI assistance was carefully reviewed, modified, and validated by the authors, and all experimental results were generated by author-executed code and independently verified through repeated runs,
cross-validation, and ablation studies. No generative AI tools were used to generate data, labels, or experimental results, nor to make scientific or
methodological decisions.

\bibliographystyle{unsrt}  
\bibliography{references}  

\appendix

\onecolumn
\begin{longtable}{p{3.4cm} p{10.3cm}}
\caption{Summary of the Hadith-inspired trust axes and engineered features with explicit, code-level computation definitions.}
\label{tab:hadith-feature-computation} \\
\toprule
\textbf{Axis} & \textbf{Feature computation (derived from implementation)} \\
\midrule
\endfirsthead

\multicolumn{2}{c}{\textit{Table \thetable\ continued}} \\
\toprule
\textbf{Axis} & \textbf{Feature computation (derived from implementation)} \\
\midrule
\endhead

\midrule
\multicolumn{2}{r}{\textit{Continued on next page}} \\
\endfoot

\bottomrule
\endlastfoot

\textbf{Integrity / Stability} (ʿAdālah)
&
\begin{itemize} 
\item \textbf{Active days}: Number of distinct calendar dates with at least one historical event prior to the window start.
\item \textbf{Total events}: Count of all user events before the window.
\item \textbf{Account age}: Number of days between the user’s earliest historical event and the window end timestamp.
\item \textbf{Daily variability}: Standard deviation of historical per-day event counts.
\item \textbf{Activity density}: Historical total events divided by the number of historical active days.
\end{itemize}
\\

\textbf{Precision / Hygiene} (Ḍabṭ)
&
\begin{itemize} 
\item \textbf{Login success rate}: Successful login/authentication events divided by total login/authentication attempts within the window.
\item \textbf{Failure delta}: Difference between the login failure rate in the window and the historical login failure rate.
\item \textbf{Burstiness}: Maximum number of events occurring within any single minute of the window.
\item \textbf{Out-of-hours fraction}: Proportion of window events occurring outside the historical 2.5--97.5 percentile activity-hour range.
\item \textbf{Timing entropy}: Shannon entropy of log-binned inter-event time gaps within the window.
\item \textbf{Path divergence}: Fraction of unique resource paths in the window that were never observed in historical activity.
\item \textbf{Sensitive-action ratio}: Proportion of window events whose event type or resource path contains sensitive-action keywords.
\end{itemize}
\\

\textbf{Chain Continuity / Context} (Isnād)
&
\begin{itemize} 
\item \textbf{IP consistency}: Inverse of the number of distinct IP addresses observed in the window.
\item \textbf{Primary IP match}: Fraction of window events whose IP matches the user’s most frequent historical IP.
\item \textbf{Subnet match}: Fraction of window IPs belonging to the user’s dominant historical /24 subnet.
\item \textbf{Geo-impossible IP switches}: Rate of IP changes occurring within less than five minutes.
\item \textbf{Session discontinuity}: Fraction of consecutive inter-event gaps within the window that exceed one hour.
\item \textbf{New IP rate}: Proportion of window IP addresses that have never appeared in the user’s history.
\end{itemize}
\\

\textbf{Reputation} (Jarḥ wa-Taʿdīl)
&
\begin{itemize} 
\item \textbf{History duration}: Total number of days spanned by the user’s historical activity.
\item \textbf{Trust ratio}: Ratio of successful historical login/authentication events to total authentication attempts.
\item \textbf{Penalty rate}: Fraction of historical events classified as failures.
\item \textbf{Failure trend}: Difference between late-history and early-history authentication failure rates.
\end{itemize}
\\

\textbf{Anomaly Evidence} (Shudhūdh / ʿIllah)
&
\begin{itemize} 
\item \textbf{Event-type KL divergence}: Kullback--Leibler divergence between window and historical event-type distributions.
\item \textbf{Hour-of-day KL divergence}: KL divergence between window and historical hour-of-day activity distributions.
\item \textbf{Path novelty}: Fraction of unique resource paths in the window that are absent from historical activity.
\item \textbf{Event-type L2 distance}: Euclidean distance between window and historical event-type frequency vectors.
\end{itemize}
\\

\multicolumn{2}{l}{\textbf{Temporal Extensions (Sequence-Aware Trust Dynamics)}} \\
\midrule

\textbf{Stability Dynamics} (Temporal ʿAdālah)
&
\begin{itemize} 
\item \textbf{Behavioral drift}: Jensen--Shannon divergence between event-type distributions of consecutive windows.
\item \textbf{Sequence entropy}: Shannon entropy of event-to-event transition probabilities within the window.
\item \textbf{Rare transition frequency}: Fraction of event transitions whose historical probability falls below a rarity threshold.
\item \textbf{Run-length anomaly}: Ratio between the longest run of identical consecutive events in the window and the expected historical run length.
\item \textbf{N-gram transition anomaly}: KL divergence between window and historical $n$-gram event-sequence distributions.
\end{itemize}
\\

\textbf{Precision Dynamics} (Temporal Ḍabṭ)
&
\begin{itemize} 
\item \textbf{Timing regularity shift}: Absolute difference in mean inter-event time between the current and previous window.
\item \textbf{Day-of-week divergence}: KL divergence between window and historical weekday activity distributions.
\item \textbf{Activity rate drift}: Difference in average event rate between consecutive windows.
\item \textbf{Authentication failure trend}: Change in authentication failure rate between consecutive windows.
\item \textbf{Cumulative suspicious activity}: Running accumulation of windows exhibiting suspicious behavioral indicators.
\end{itemize}
\\

\textbf{Continuity Dynamics} (Temporal Isnād)
&
\begin{itemize} 
\item \textbf{Subnet drift}: Indicator of change in dominant IP subnet between consecutive windows.
\item \textbf{IP switch rate}: Frequency of IP address changes normalized by window duration.
\item \textbf{Device transition anomaly}: KL divergence between window and historical device-transition distributions.
\end{itemize}
\\

\textbf{Reputation Dynamics} (Temporal Reputation)
&
\begin{itemize}
\item \textbf{Risk acceleration}: Second-order difference of the inferred risk score across consecutive windows.
\item \textbf{Transition-distribution divergence}: KL divergence between window-level and historical event-transition matrices.
\item \textbf{Activity autocorrelation}: Temporal autocorrelation of activity rates across consecutive windows.
\end{itemize}
\\

\end{longtable}

\section{Complete Feature Formalization (42 Features)}
This appendix provides a complete mathematical formulation of the 42 engineered features summarized in Table~\ref{tab:hadith-feature-computation}.

\subsection{Notation}
Let:
\begin{itemize}
    \item $U$ denote a user
    \item $H_U = \{e_1,\dots,e_n\}$ denote all historical events of user $U$
    \item $W_t = \{w_1,\dots,w_m\}$ denote the $t$-th sliding window
    \item $t(e)$ denote the timestamp of event $e$
    \item $\mathrm{date}(\cdot)$ extract the calendar date from a timestamp
    \item $h(e)$ extract the hour-of-day from a timestamp
    \item $\Delta t_i = t(w_{i+1}) - t(w_i)$ denote consecutive inter-event gaps
    \item $\mathcal{E}$ denote the set of event types
    \item $\mathcal{P}$ denote the set of resource paths
    \item $\mathcal{I}$ denote the set of IP addresses
    \item $\mathcal{I}_{W_t}$, $\mathcal{P}_{W_t}$ denote IPs and paths observed in window $W_t$
    \item $\mathcal{I}_{H_U}$, $\mathcal{P}_{H_U}$ denote IPs and paths observed in history $H_U$
    \item $\text{IP}(e)$ denote the IP address of event $e$
    \item $\text{subnet}(e)$ denote the /24 subnet of the IP address of event $e$
    \item $\text{IP}^*_U$ denote the most frequent historical IP address of user $U$
    \item $\text{subnet}^*_U$ denote the most frequent historical subnet of user $U$
    \item $\text{Login}(\cdot)$ denote authentication attempts
    \item $\text{Success}(\cdot)$ denote successful authentication events
    \item $\text{Fail}(\cdot)$ denote failed authentication events
    \item $\text{Sensitive}(\cdot)$ denote events involving sensitive actions
    \item $P_X(\cdot)$ denote an empirical probability distribution estimated from set $X$
    \item $\mathbf{v}_X$ denote a normalized event-type frequency vector for set $X$
    \item $P(i \rightarrow j)$ denote the probability of event-type transition $i \rightarrow j$
    \item $T_X$ denote the event-transition probability matrix estimated from $X$
    \item $P_{\text{dow}}^X$ denote the day-of-week distribution estimated from $X$
    \item $P_{\text{device}}^X$ denote the device-type distribution estimated from $X$
    \item $Q_{2.5}, Q_{97.5}$ denote the 2.5th and 97.5th percentiles of historical activity hours
    \item $\tau$ denote a rarity threshold for transitions
    \item $R_t$ denote the aggregate risk score at window $t$
    \item $\text{Rate}(W_t)=|W_t|/\Delta T_t$ denote the activity rate in window $W_t$
    \item $\epsilon$ denote a small constant to avoid division by zero
\end{itemize}

\subsection{Integrity / Stability (ʿAdālah)}

\begin{equation}
\text{ActiveDays}(U)=\left|\{\mathrm{date}(t(e)) : e\in H_U\}\right|
\end{equation}

\begin{equation}
\text{TotalEvents}(U)=|H_U|
\end{equation}

\begin{equation}
\text{AccountAge}(U,W_t)=\max_{w\in W_t}t(w)-\min_{e\in H_U}t(e)
\end{equation}

\begin{equation}
\text{DailyStd}(U)=\operatorname{StdDev}\left(\{|H_U(d)|\}_d\right)
\end{equation}

\begin{equation}
\text{EventsPerDay}(U)=\frac{|H_U|}{\text{ActiveDays}(U)+\epsilon}
\end{equation}

\subsection{Precision / Hygiene (Ḍabṭ)}

\begin{equation}
\text{LoginSuccessRate}(W_t)=
\frac{|\text{Success}(W_t)|}{|\text{Login}(W_t)|+\epsilon}
\end{equation}

\begin{equation}
\Delta\text{FailRate}=
\text{FailRate}(W_t)-\text{FailRate}(H_U)
\end{equation}

\begin{equation}
\text{Burstiness}(W_t)=
\max_m\left|\left\{w\in W_t:
\left\lfloor\frac{t(w)}{60}\right\rfloor=m\right\}\right|
\end{equation}

\begin{equation}
\text{OutOfHours}(W_t)=
\frac{|\{w\in W_t:h(w)\notin[Q_{2.5},Q_{97.5}]\}|}{|W_t|}
\end{equation}

\begin{equation}
\text{TimingEntropy}(W_t)=
-\sum_i p_i\log p_i
\end{equation}

\begin{equation}
\text{PathDivergence}(W_t,H_U)=
\frac{|\mathcal{P}_{W_t}\setminus\mathcal{P}_{H_U}|}
{|\mathcal{P}_{W_t}|+\epsilon}
\end{equation}

\begin{equation}
\text{SensitiveRatio}(W_t)=
\frac{|\text{Sensitive}(W_t)|}{|W_t|}
\end{equation}

\subsection{Chain Continuity / Context (Isnād)}

\begin{equation}
\text{IPConsistency}(W_t)=
\frac{1}{|\mathcal{I}_{W_t}|+\epsilon}
\end{equation}

\begin{equation}
\text{PrimaryIPMatch}(W_t)=
\frac{|\{w\in W_t:\text{IP}(w)=\text{IP}^*_U\}|}{|W_t|}
\end{equation}

\begin{equation}
\text{SubnetMatch}(W_t)=
\frac{|\{w\in W_t:\text{subnet}(w)=\text{subnet}^*_U\}|}{|W_t|}
\end{equation}

\begin{equation}
\text{GeoImpossible}(W_t)=
\frac{|\{(w_i,w_{i+1}):\Delta t_i<300\land
\text{IP}(w_i)\neq\text{IP}(w_{i+1})\}|}{|W_t|}
\end{equation}

\begin{equation}
\text{SessionDiscontinuity}(W_t)=
\frac{|\{\Delta t_i>3600\}|}{|W_t|}
\end{equation}

\begin{equation}
\text{NewIPRate}(W_t,H_U)=
\frac{|\mathcal{I}_{W_t}\setminus\mathcal{I}_{H_U}|}
{|\mathcal{I}_{W_t}|}
\end{equation}

\subsection{Reputation (Jarḥ wa-Taʿdīl)}

\begin{equation}
\text{HistoryDuration}(U)=
\max_{e\in H_U}t(e)-\min_{e\in H_U}t(e)
\end{equation}

\begin{equation}
\text{TrustRatio}(H_U)=
\frac{|\text{Success}(H_U)|}{|\text{Login}(H_U)|+\epsilon}
\end{equation}

\begin{equation}
\text{PenaltyRate}(H_U)=
\frac{|\text{Fail}(H_U)|}{|H_U|}
\end{equation}

\begin{equation}
\text{FailureTrend}(U)=
\text{FailRate}(H_U^{\text{late}})-
\text{FailRate}(H_U^{\text{early}})
\end{equation}

\subsection{Anomaly Evidence (Shudhūdh / ʿIllah)}

\begin{equation}
D_{\mathrm{KL}}(W_t||H_U)=
\sum_{e\in\mathcal{E}}
P_{W_t}(e)\log\frac{P_{W_t}(e)}{P_{H_U}(e)}
\end{equation}

\begin{equation}
D_{\mathrm{KL}}^{\text{hour}}(W_t||H_U)=
D_{\mathrm{KL}}(P_{\text{hour}}^{W_t}||P_{\text{hour}}^{H_U})
\end{equation}

\begin{equation}
\text{PathNovelty}(W_t,H_U)=
\frac{|\mathcal{P}_{W_t}\setminus\mathcal{P}_{H_U}|}
{|\mathcal{P}_{W_t}|}
\end{equation}

\begin{equation}
D_{L2}(W_t,H_U)=
\|\mathbf{v}_{W_t}-\mathbf{v}_{H_U}\|_2
\end{equation}

\subsection{Temporal Trust Dynamics (16)}

\begin{equation}
\text{BehaviorDrift}(t)=
D_{\mathrm{JS}}(P_{W_t},P_{W_{t-1}})
\end{equation}

\begin{equation}
\text{SequenceEntropy}(W_t)=
-\sum_{i,j}P(i\rightarrow j)\log P(i\rightarrow j)
\end{equation}

\begin{equation}
\text{RareTransitionRate}(W_t)=
\frac{|\{(i,j):P_U(i\rightarrow j)<\tau\}|}{|W_t|}
\end{equation}

\begin{equation}
\text{RunLengthAnomaly}(W_t)=
\frac{\max\text{RunLength}(W_t)}
{\mathbb{E}[\text{RunLength}(H_U)]}
\end{equation}

\begin{equation}
\text{NGramAnomaly}(W_t)=
D_{\mathrm{KL}}(P_{n\text{-gram}}^{W_t}||
P_{n\text{-gram}}^{H_U})
\end{equation}

\begin{equation}
\text{TimingShift}(t)=
|\mu_{\Delta t}^{W_t}-\mu_{\Delta t}^{W_{t-1}}|
\end{equation}

\begin{equation}
\text{WeekdayDivergence}(W_t)=
D_{\mathrm{KL}}(P_{\text{dow}}^{W_t}||
P_{\text{dow}}^{H_U})
\end{equation}

\begin{equation}
\text{RateDrift}(t)=
\text{Rate}(W_t)-\text{Rate}(W_{t-1})
\end{equation}

\begin{equation}
\text{FailureTrend}(t)=
\text{FailRate}(W_t)-\text{FailRate}(W_{t-1})
\end{equation}

\begin{equation}
\text{CumulativeSuspicious}(t)=
\sum_{k=1}^{t}\mathbb{I}[\text{Suspicious}(W_k)]
\end{equation}

\begin{equation}
\text{SubnetDrift}(t)=
\mathbb{I}[\text{subnet}_t\neq\text{subnet}_{t-1}]
\end{equation}

\begin{equation}
\text{IPSwitchRate}(t)=
\frac{|\Delta\text{IP}|}{\Delta t}
\end{equation}

\begin{equation}
\text{DeviceTransitionAnomaly}(W_t)=
D_{\mathrm{KL}}(P_{\text{device}}^{W_t}||
P_{\text{device}}^{H_U})
\end{equation}

\begin{equation}
\text{RiskAcceleration}(t)=
R_t-2R_{t-1}+R_{t-2}
\end{equation}

\begin{equation}
\text{TransitionKL}(t)=
D_{\mathrm{KL}}(T_{W_t}||T_{H_U})
\end{equation}

\begin{equation}
\text{Autocorrelation}=
\operatorname{corr}(\text{Rate}(W_t),\text{Rate}(W_{t-1}))
\end{equation}

\end{document}